\pdfoutput=1
\documentclass[letterpaper, 10 pt, conference]{ieeeconf}  % Comment this line out if you need a4paper

\IEEEoverridecommandlockouts                              % This command is only needed if 
\usepackage{cite}
\usepackage{amsmath,amssymb,amsfonts}
\usepackage{algorithmic}
\usepackage{graphicx}
\usepackage{textcomp}
\usepackage{xcolor}
\usepackage{booktabs}
\usepackage{multicol, blindtext}
\usepackage[bookmarks=false]{hyperref}
\usepackage{caption}
\usepackage{multirow}
\usepackage{siunitx}
\usepackage{etoolbox}
\usepackage{array}
\usepackage{gensymb}
\usepackage{subfigure}
\usepackage{textcomp}

\usepackage{multicol, blindtext}
\usepackage{authblk}
\usepackage{hyperref}
\usepackage{siunitx}
\usepackage{etoolbox}
\usepackage[labelsep=period]{caption}
\usepackage[style=ieee]{}
\usepackage{siunitx}
\usepackage{float} 
\usepackage{flushend}

\title{\LARGE\bf Visual Localization for Autonomous Driving: Mapping the Accurate Location in the City Maze}

\author{Dongfang Liu, Yiming Cui, Xiaolei Guo, Wei Ding, Baijian Yang, and Yingjie Chen% <-this % stops a space
\thanks{D. Liu, X. Guo, W. Ding, B. Yang and Y. Chen are with Purdue University. 
    {\tt\small\{liu2538,guo579,ding242,byang and victorchen\}@purdue.edu}}\\
\thanks{Y. Cui is with University of Florida.
        {\tt\small cuiyiming@ufl.edu}}%
}

\begin{document}

\maketitle
\thispagestyle{empty}
\pagestyle{empty}

\setlength{\textfloatsep}{1pt}% Remove
%%%%%%%%%%%%%%%%%%%%%%%%%%%%%%%%%%%%%%%%%%%%%%%%%%%%%%%%%%%%%%%%%%%%%%%%%%%%%%%%
\begin{abstract}
Accurate localization is a foundational capacity, required for autonomous vehicles to accomplish other tasks such as navigation or path planning. It is a common practice for vehicles to use GPS to acquire location information. However, the application of GPS can result in severe challenges when vehicles run within the inner city where different kinds of structures may shadow the GPS signal and lead to inaccurate location results. To address the localization challenges of urban settings, we propose a novel feature voting technique for visual localization. Different from the conventional front-view-based method, our approach employs views from three directions (front, left, and right) and thus significantly improves the robustness of location prediction. In our work, we craft the proposed feature voting method into three state-of-the-art visual localization networks and modify their architectures properly so that they can be applied for vehicular operation. Extensive field test results indicate that our approach can predict location robustly even in challenging inner-city settings. Our research sheds light on using the visual localization approach to help autonomous vehicles to find accurate location information in a city maze, within a desirable time constraint.
\end{abstract}

\section{Introduction}
\begin{figure}[!ht]
  \centering
    \subfigure[]{
        \includegraphics[width=8.5cm]{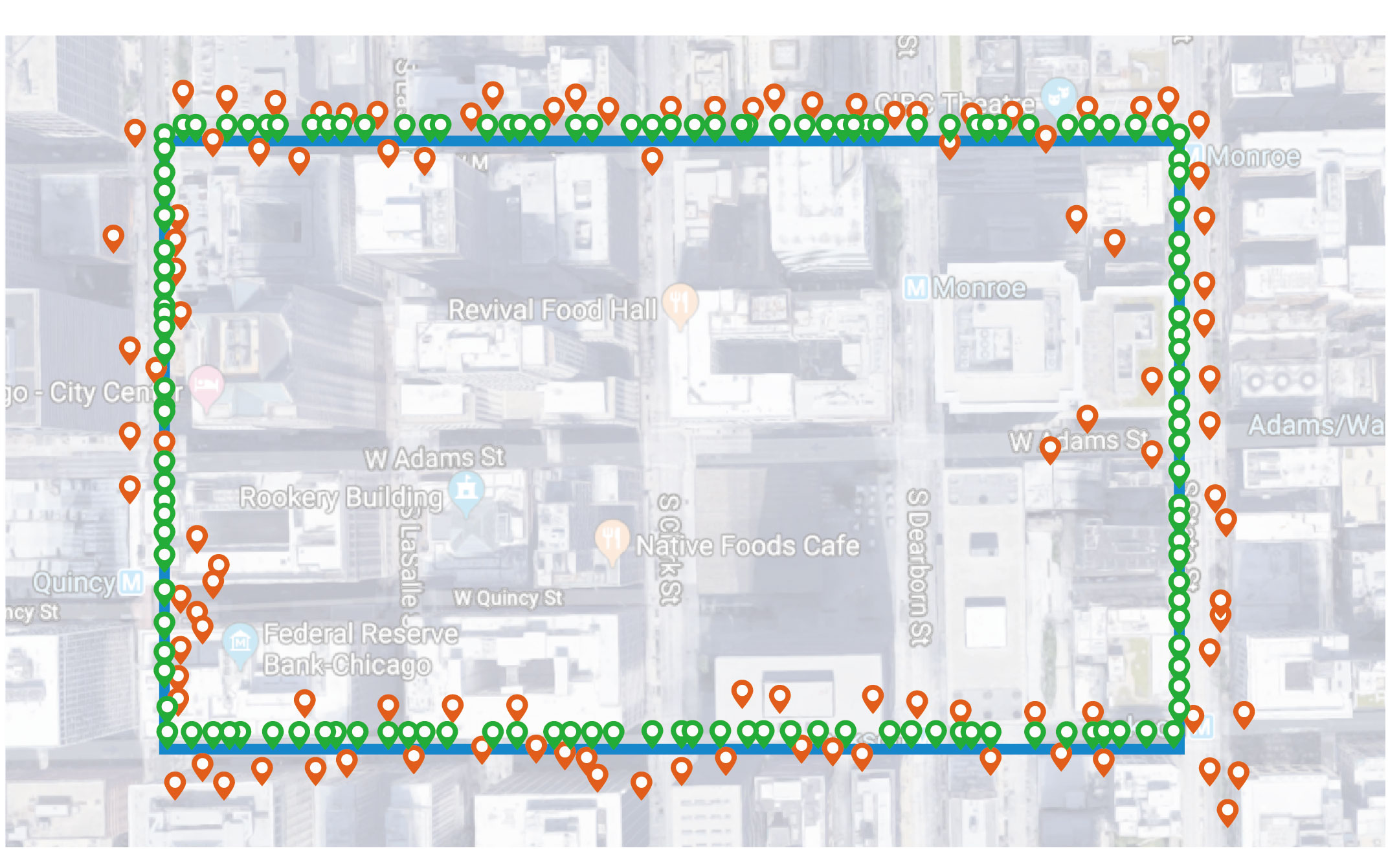}
    }
     \subfigure[]{
     \includegraphics[width=8.5cm]{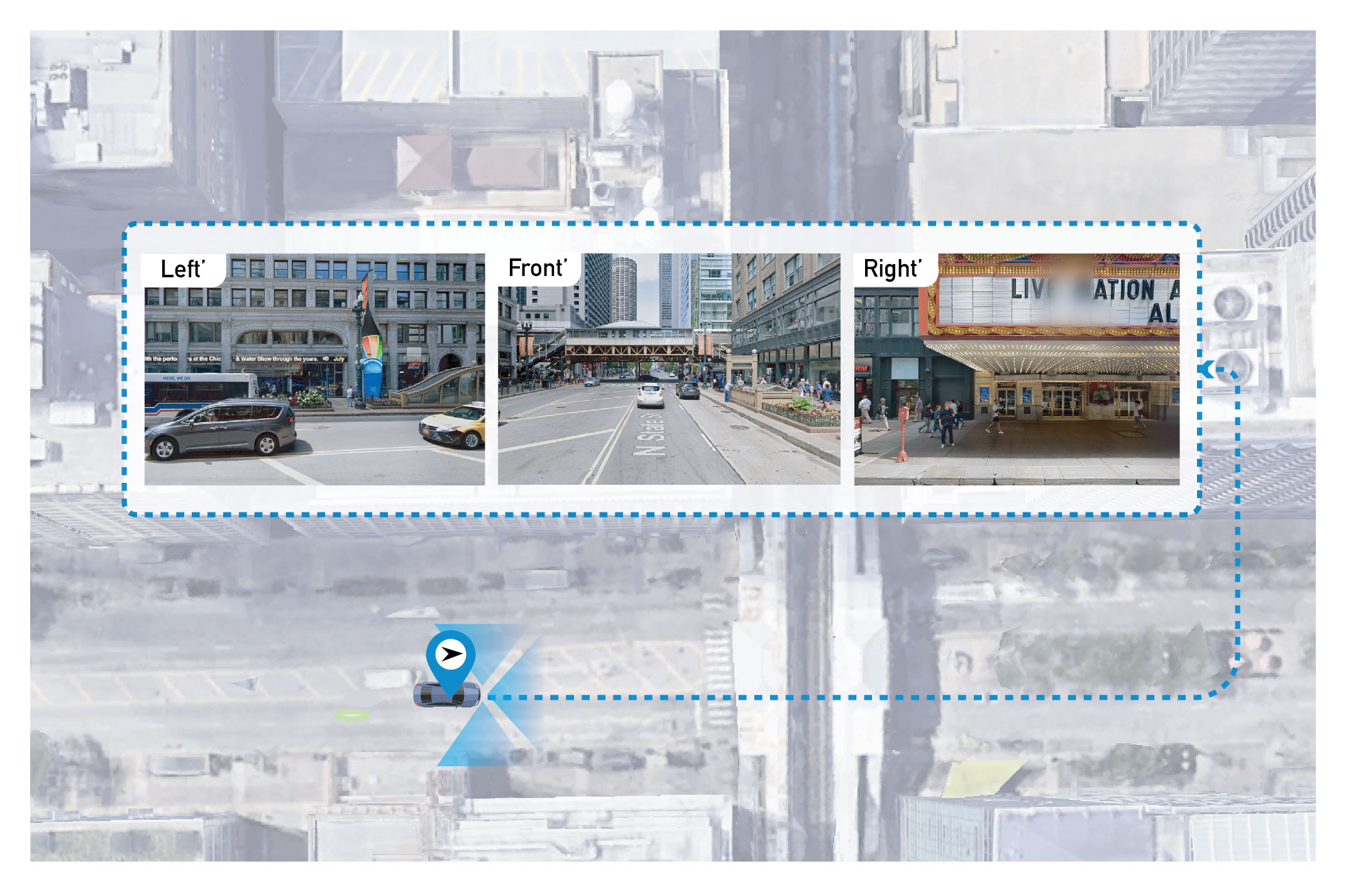}
    }
  \caption{In (a), the blue line indicates the vehicular trajectory, the red icons are the GPS signals, and the green icons are the predicted locations by our visual localization network. For a clear visualization, GPS and location prediction are shown here for every second. In (b), we demonstrate using the  front, left, and right views for location prediction.}
  \label{Figure 1}
\end{figure}

\indent Accurate localization is a foundational requirement for autonomous vehicles to accomplish other tasks such as navigation or path planning\cite{9206716}\cite{LIU20201}. In common practice, vehicles utilize GPS data to obtain location information. The stability of this approach, however, relies purely on instant access to satellite signals, which could be frequently influenced by environmental elements such as buildings, trees, or weather \cite{salarian2018improved}\cite{liu2019virtual}\cite{9207265}. Specifically, it is extremely challenging to obtain accurate localization using GPS for vehicles that move on streets surrounded by dense skyscrapers in inner-city regions \cite{sattler2018benchmarking}. The compromised GPS signal can inaccurately place the moving vehicle in a location that is far from the ground truth (as shown in Fig. \ref{Figure 1}(a)). Such erroneous localization is unacceptable for the operation of autonomous vehicles.\\
\indent Therefore, significant research effort has been directed at finding an alternative solution to improve localization. Recent computer vision advances have made it possible for visual localization to search pre-stored datasets with sufficient reliability \cite{guan2013device}\cite{li2012interactive}\cite{yan2017supervised}. The motivation behind our work is to develop a visual localization method that assists an autonomous vehicle in finding accurate location information in dense urban settings where GPS signals are partially or completely shadowed.\\
\begin{figure*}[!ht]
  \centering
\includegraphics[width=17cm]{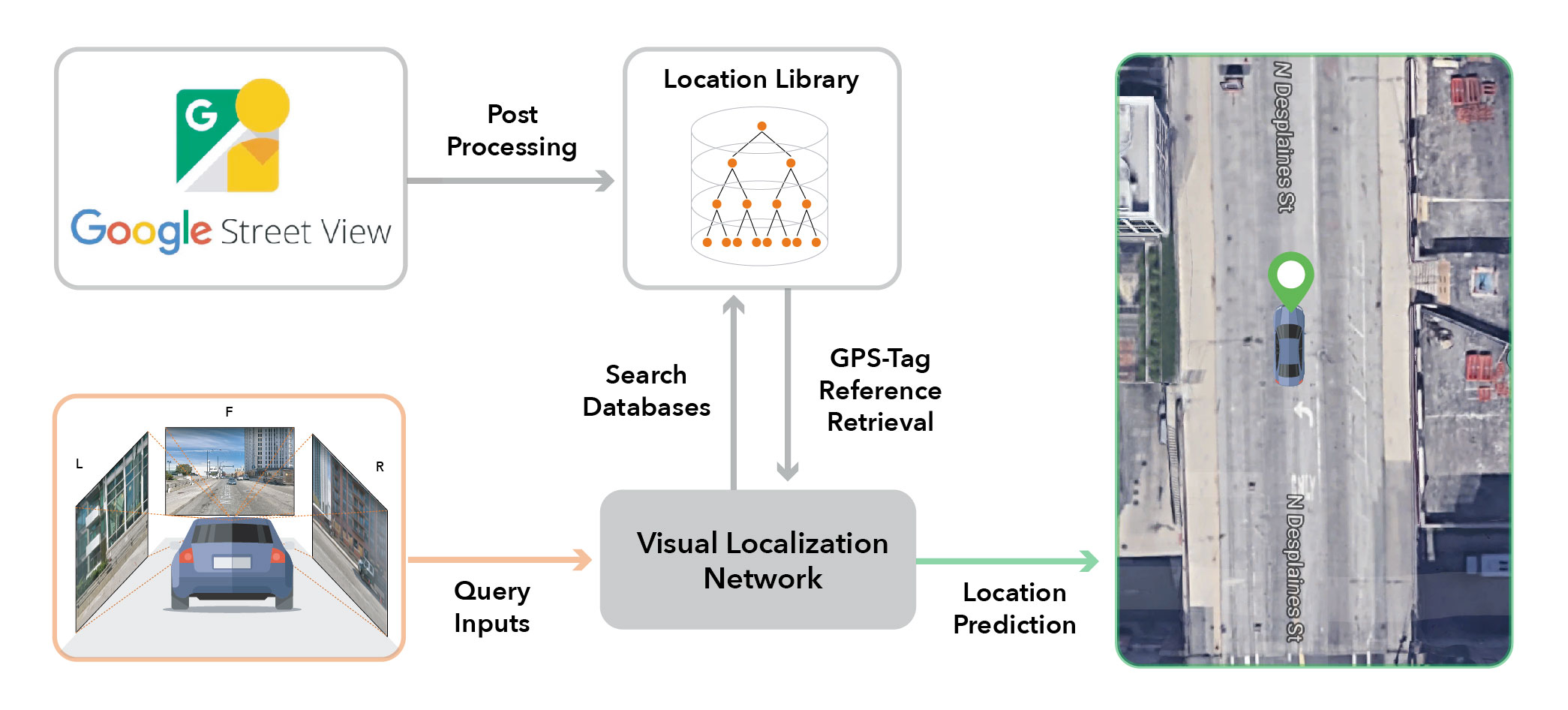}
  \caption{The framework of the proposed method. We use Google Street View to collect front, left and right view images along the driving trajectory. The collected images are post-processed to construct a location library. In inference, the visual localization network, using the proposed feature voting method, takes the query inputs and searches the location library by matching the local features of the GPS-tagged reference images. The GPS tag from the references with the lowest voting cost is retrieved as the location prediction.}
  \label{Figure 2}
\end{figure*}
\indent Visual localization relies on identifying local features and searching over a pre-stored GPS-tagged image database to find a matching reference image and predict the current location \cite{agarwal2015metric}\cite{luo2019image}\cite{salarian2018improved}. Existing work on visual localization generally stores and collects only front-view image data for matching \cite{salarian2018improved}\cite{agarwal2015metric}\cite{sarlin2019coarse}\cite{liu2019stochastic}. If local features between the query image and the reference image from the database closely match, we retrieve that specific reference image and use its GPS tag to predict the location of the query input. However, with the increase of database size, front-view inputs from an urban setting can be homogeneous, with too much architectural similarities for the matching algorithm to work robustly \cite{salarian2018improved}\cite{zamir2014image}. To address this issue, \cite{salarian2018improved}\cite{hao20123d}\cite{Brosh_2019_CVPR_Workshops}\cite{zhang2011image} consider global features such as color histogram, GIST, and GPS coordinates to determine the final location inference. However, these approaches increase computation load and slow the speed which means less power and speed are available for vehicular operation.\\
\indent In this paper, we introduce a novel approach for visual localization which can accurately locate vehicles in inner-city areas where GPS signals are shadowed and road conditions are complex. Instead of using global features, we use front, left, and right views to triangulate vehicle location (as shown in Fig. \ref{Figure 1}(b)). Since the left and right views generally change faster than the front view when a vehicle moves forward, left and right images offer more  useful information with which to differentiate close references from the pre-stored database. We leverage Google Street View to build the pre-stored, GPS-tagged image database. Since Google Street View has spherical 360$^{\circ}$ views, we can easily access the front, left, and right views from a single location. Compared to the conventional front-view approach, our method improves location accuracy significantly.\\
\indent The major contributions of our work are threefold: 1. We design a pipeline to implement a three-directional-view-based visual localization system for vehicles. Using three-directional views to map the GPS-tagged reference significantly improves location accuracy for vehicular operation. 2. We implement an automated pipeline to collect, annotate, and manage GPS-tag data for visual localization. We leverage the cheap data source from Google Street View which is easy to access. Our method reduces the cost of GPS-tag data collection, annotation, and management. 3. Finally, we introduce a GPS-tag location dataset which helps other researchers to explore and  investigate visual localization for autonomous vehicles.
\section{RELATED WORK}
\indent The localization of vehicles primarily uses GPS data. The adequacy of this approach depends mostly on satisfactory access to satellite signals. In practice, GPS information is usually reliable when the device has a clear view of the sky to get a signal from at least four satellites \cite{salarian2018improved}. However, it is difficult to obtain accurate localization using GPS in a dense urban area where road conditions are complex and GPS signals are frequently blocked by skyscrapers.\\
\indent Alternatively, other works have used visual localization which compares local features from front-view images to predict locations \cite{babenko2014neural}\cite{toft2018semantic}\cite{gordo2016deep}\cite{gordo2017end}. Although the local feature matching approach is proven to be powerful for visual localization, its performance degrades when the homogeneousness of key-point features occurs substantially \cite{sarlin2019coarse}\cite{liu2019stochastic}\cite{arandjelovic2016netvlad}. One typical example can be frequently observed in urban areas where man-made structures have such architectural similarity that descriptors become difficult to differentiate among similar images taken from one direction \cite{salarian2018improved}\cite{majdik2013mav}. To resolve the above challenges, \cite{vishal2015accurate}\cite{roshan2014gps}\cite{zamir2010accurate}\cite{agarwal2015metric} employ different types of global features, such as color histogram, GIST, GPS coordinates, and odometry data to predict the final inference. However, the gain from using global features comes at the price of longer matching times and higher memory consumption, and the results are still not robust enough \cite{sarlin2019coarse}.\\
\indent Different from existing methods, our approach only uses visual information from camera sensors to predict locations. There are existing methods from \cite{meilland2015dense}
 \cite{bresson2019urban}\cite{Sattler_2018_CVPR}\cite{stenborg2018long} that also leverage camera images to locate the mobile agent. However, \cite{meilland2015dense}
 \cite{bresson2019urban} use panorama or spherical views for feature storage and matching which requires a large computation load; \cite{stenborg2018long} 
 \cite{Sattler_2018_CVPR} require expensive high dimensional image data, such as 3D point maps and 6 degree-of-freedom (6DOF) camera inputs for training and prediction. The inference speed for the aforementioned methods is slow and cannot be used for vehicular operation. To address these limitations, we use RGB images from three directions (front, left and right views) to quickly triangulate the location and robustly predict the location inference. A complimentary feature voting technique is crafted into a visual localization network to determine the contribution of each view direction in location prediction. Based on experimental results, our method is applicable for real-time vehicle operation.
\section{Method and Experiments}
In this section, we elaborate on our method for our experiments. Fig. \ref{Figure 2} demonstrates the framework of our work. We leverage Google Street View to construct the location library. To predict location, the visual localization network takes query inputs and searches over database from location library. The reference which has the smallest voting cost for image matching to the query inputs is then retrieved. The location prediction is based on the GPS tag of the retrieved reference. The details are introduced in the following sections.
\subsection{Location Library}
\subsubsection{Automated data collection}
Following the paradigm of visual localization, we acquire location information beforehand based on Google Street View. We develop an automated image collection method based on Google Street View with which to build our location library. Google Street View provides high-quality interactive 360$\degree$ panoramas labeled with GPS coordinates. Covering almost all street information for major cities in the world, Google Street View becomes an effective tool for obtaining street images in complicated urban areas.
\begin{figure}[!ht]
  \centering
\includegraphics[width=6cm]{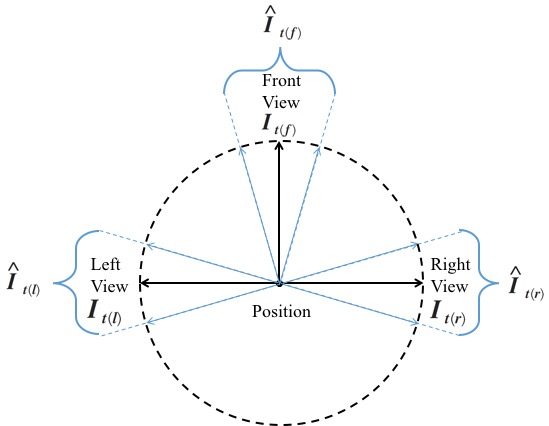}
  \caption{Illustration of camera orientation.}
  \label{Figure 3}
\end{figure}
\subsubsection{Post processing}
\indent The process of Google Street View image collection can be divided into three steps: 1. Interacting with Google Street View, we first obtain front-view information (GPS coordinates, camera orientation, and image ${I}_{t(f)}$ where $t = 1, 2, 3, \dots$) for the selected street through parsing URLs at each position on the street. 2. based on information obtained in step 1, for each street position, we generate two other URLs by resetting the camera orientation parameters to get street views in different directions. To be specific, as shown in Fig. \ref{Figure 3}, we rotate the camera orientation $90\degree$ counterclockwise and clockwise to get the left-view image ${I}_{t(l)}$ and right-view image ${I}_{t(r)}$. Based on the positions of ${I}_{t(f)}, {I}_{t(l)}$ and ${I}_{t(r)}$, we rotate the camera orientations another $\pm 5\degree$ ($0\degree$ not included) randomly at front-view, left-view and right-view to get their nearby images $\hat{I}_{t(f)}, \hat{I}_{t(l)}$ and $\hat{I}_{t(r)}$, respectively. 3. Finally, we extract and save street view images for selected streets by visiting the three Google Street View URLs at each position. All the aforementioned steps are implemented in an automated pipeline. Situations like occlusions, different weather, and light conditions are not taken into account but will be considered in future work to improve the robustness of the proposed method. After data collection, the front, left, and right views $\left({I}_{t(f)}, {I}_{t(l)}, {I}_{t(r)}\right)$ are used to build the location library, while $({I}_{t(f)}, \hat{I}_{t(f)})$, $({I}_{t(l)}, \hat{I}_{t(l)})$ and $({I}_{t(r)}, \hat{I}_{t(r)})$ are used as image pairs respectively for training feature descriptors. We use VGG16 architecture \cite{simonyan2014very} as the feature extraction network to process the collected GPS-tagged images from the previous step. In our experiment, we remove the FC layers and only use convolution layers to perform the feature extraction. The outcome of the last convolution layers is further processed by NetVLAD pooling \cite{arandjelovic2016netvlad} in order to obtain a $1 \times 256$ feature vector. Then, the obtained feature vectors are \textit{L2} normalized. Once we process all the collected images, we aggregate the local feature vectors and organize them into a $k$-means reference tree \cite{muja2009fast}. The location library, where we conduct feature matching and reference retrieval, consists of 7,456 geo-locations and the reference tree. 
% \textcolor{green}{.}
% \textcolor{red}{We conduct feature matching and reference retrieval over our location library.}

\subsection{Feature Voting on Reference Retrieval}
\begin{figure}[!ht]
  \centering
\includegraphics[width=8.5cm]{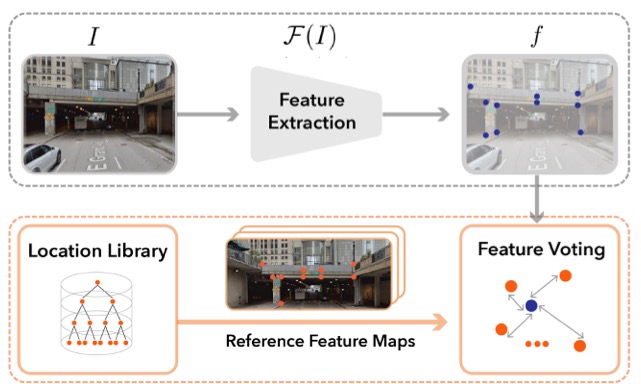}
  \caption{The feature extraction and feature voting process.}
  \label{Figure 4}
\end{figure}

\indent In our experiments, we craft three state-of-the-art visual localization networks (NetVLAD\cite{arandjelovic2016netvlad}, SARE\cite{liu2019stochastic}, and D2-Net \cite{dusmanu2019d2}) using our feature voting method. In order to achieve a faster inference speed, we replace VGG16\cite{simonyan2014very} from the original architectures with MobileNet \cite{howard2017mobilenets} as the feature extraction network $\mathcal{F}$.  
The query image $I_{q}$ is processed by $\mathcal{F}(I_{q})$ which generates feature maps $f_{q}$ (as shown in the gray section of  Fig. \ref{Figure 4}). Feature maps $f_{q}$ include all the local feature points on the image.
\\
\indent Then, we use a voting operation $\mathcal{V}$ to compute the voting cost $\mathcal{S}$:
\begin{equation}
    \mathcal{S} =\mathcal{V}\left(f_{q},f_{r}\right) = \sum_{i=1}^{C}\parallel f^{i}_{q} - \left(f^{i}_{r}\right)\parallel
    \label{Equation 1},
\end{equation}
where $C$ is the number of local feature points on the image, $f^{i}_{q}$ and $f^{i}_{r}$ are the $i^{th}$ feature points on the query and reference feature maps respectively, and $\parallel.\parallel$ calculates the distance between the two features. The feature distance indicates the similarity between the reference feature and the corresponding query feature (as shown in the orange section of Fig. \ref{Figure 4}). A smaller value for the voting cost $\mathcal{S}$ indicates the matching features have a higher similarity and vice versa.\\
\indent The reference which has the lowest voting cost $\mathcal{S}$ is retrieved from the location library. For our approach, the image retrieval considers voting for three directions, so we design an adaptive weight $\alpha$ to assist in the reference retrieval from the location library. Hence, our total voting cost is defined as: 
\begin{equation}
    \mathcal{S}_{Total} = \alpha \mathcal{S}_{F}+\frac{(1-\alpha)}{2}(\mathcal{S}_{L}+\mathcal{S}_{R}),
    \label{Equation 2}
\end{equation}
where $S_{F}$, $S_{L}$, and $S_{R}$ are voting costs for front, left, and right views. $\alpha$ is the adaptive weight which determines the contribution of each direction to the final retrieval decision. $ \mathcal{S}_{Total}$ is the total cost for an image set (including front, left, and right views). Our goal is to retrieve  the GPS-tagged reference with the smallest $ \mathcal{S}_{Total}$. Our visual localization method is summarized in Algorithm 1.
\begin{table}
\scriptsize
\centering
\caption*{}
  \begin{tabular} {c}
%   {\textwidth}{>{\hspace{-6pt}}ll<{\hspace{-4pt}}}
    \toprule
    \textbf{Algorithm 1} Image Set Retrieval Process.\\
    \midrule
    1: \textbf{input}: frame{$\{I_{t(f)},I_{t(l)},I_{t(r)}\}$}  \hfill $\triangleleft$ Input image sets, $t=1, \dots, \infty$\\
    2: \textbf{while not} find a matching reference \textbf{do}  \hfill $\triangleleft$ Search the location library\\
    3: \hspace{0.3cm}$f_{t(f)},f_{t(l)},f_{t(r)} =\mathcal{F}\left(I_{t(f)},I_{t(l)},I_{t(r)}\right)$  \hfill $\triangleleft$ Extract local features\\
    4: \hspace{0.3cm}$S_{t(f)},S_{t(l)},S_{t(r)} =\mathcal{V}\left(f_{t(f)},f_{t(l)},f_{t(r)}\right)$  \hfill $\triangleleft$ Each voting cost\\
    5: \hspace{0.3cm}$\mathcal{S}_{t(total)} =  \alpha\mathcal{S}_{t(f)}+\frac{(1-\alpha)}{2}(\mathcal{S}_{t(l)}+\mathcal{S}_{t(r)})$  \hfill $\triangleleft$ Total voting cost\\
    6: \hspace{0.3cm}$B =\mathcal{A}(\mathcal{S}_{t(total)},\mathcal{L}_{GPS})$  \hfill $\triangleleft$ Add information in the buffer\\
    7: \textbf{return}: $\mathop{\min}(B)$ \hfill $\triangleleft$ Retrieve the reference with the smallest cost\\
    \bottomrule
  \end{tabular}
\end{table}
\subsection{Learning Objective}
We design a modified triplet margin ranking loss $\mathcal{L}$ \cite{mishchuk2017working} to optimize the local feature extraction. Given a pair of images $({I}, {I}^{'})$ and a corresponding feature point
$c : P \leftrightarrow P^{'}$ between them (where $P \in {I}^{}, P^{'} \in {I}^{'}$),
we want to minimize the distance of the corresponding descriptors $\hat{d}_{P}$ and $\hat{d}_{P'}$ and maximize the distance of confounding descriptors $ \hat{d}_{N_{}}$ or $\hat{d}_{N'}$ in each image. To achieve this learning objective, we define the positive descriptor distance $p(c)$ between the corresponding descriptors which is $\parallel\hat{d}_{P}-\hat{d}_{P'}\parallel_{2}.$ Similarly, the negative distance $n(c)$, which stems from the most confounding descriptor for either $\hat{d}_{N}$ or $\hat{d}_{N'}$, is defined as $ min( \parallel \hat{d}_{P}- \hat{d}_{N'} \parallel_{2}, \parallel \hat{d}_{N} - \hat{d}_{P'} \parallel_{2}).$ We use $p(c)$ and $n(c)$ to define the triplet margin ranking loss with a margin $M$ as:
\begin{equation}
m\left(c\right)=\max\left(M + p\left(c\right)^2 - n \left(c\right)^2,0\right)
\label{Equation 3}.
\end{equation}
Accordingly, our training objective can be defined as:\\
\begin{equation}
    \mathcal{L}\left(I, I^{'}\right) =  \sum_{c \in \mathcal{C}} m\left(p\left(c\right), n\left(c\right)\right)
    \label{Equation 4},
\end{equation}
where $\mathcal{C}$ is the set of all corresponding feature points between ${I}$ and ${I}^{'}$. While the loss is minimized, the most distinctive correspondences will get higher detection scores which will lead to a larger distinction between positive descriptors and negative descriptors. Training is conducted on the image pairs collected from Google Street View.

\begin{figure}[!ht]
  \centering
\includegraphics[width=8cm]{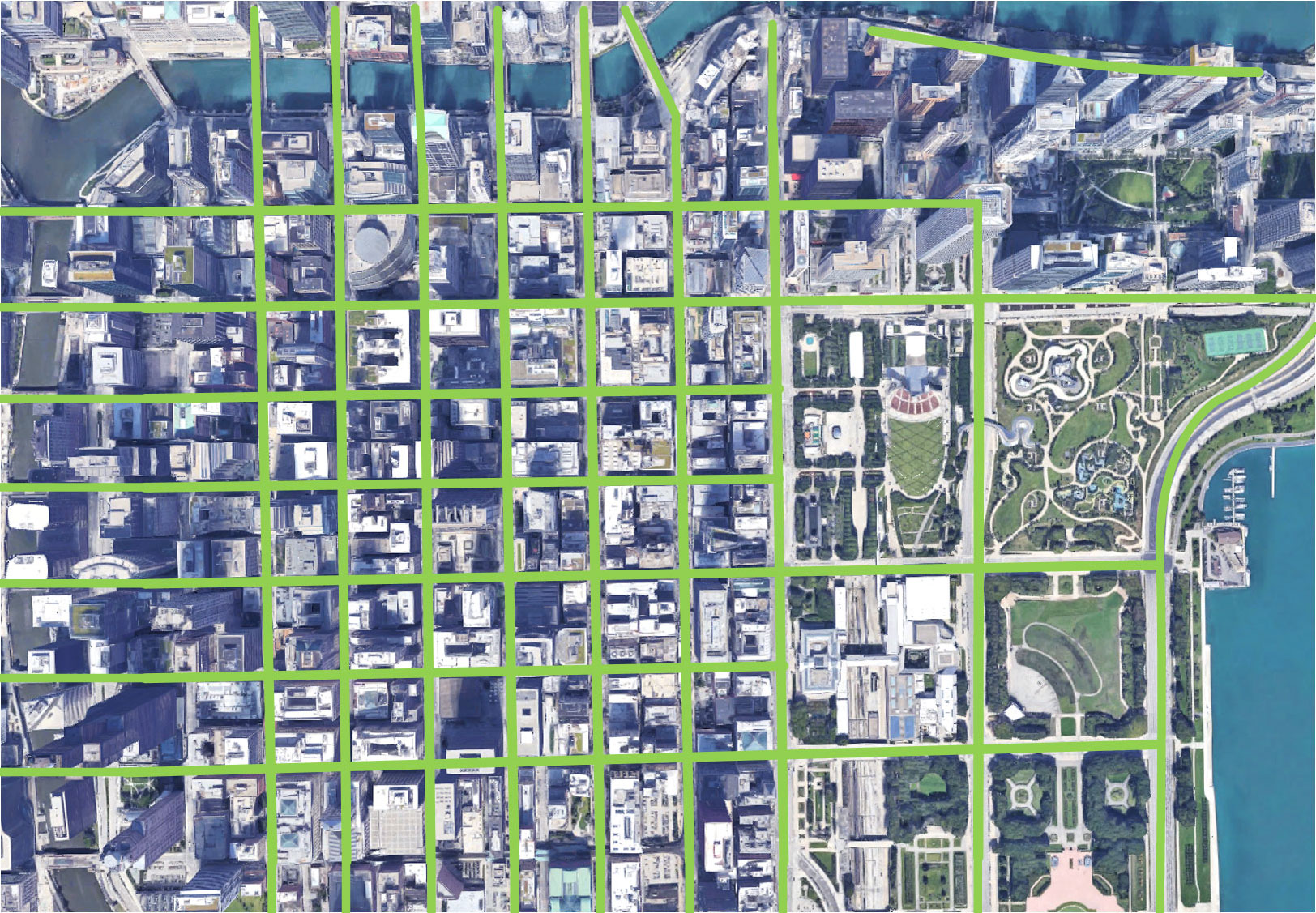}
  \caption{Aerial-view map of downtown in Chicago, IL. Green lines represent target streets whose street images have been extracted with our automated data collection method.}
  \label{Figure 5}
\end{figure}
\subsection{Dataset}
\begin{figure}[!ht]
  \centering
\includegraphics[width=7cm]{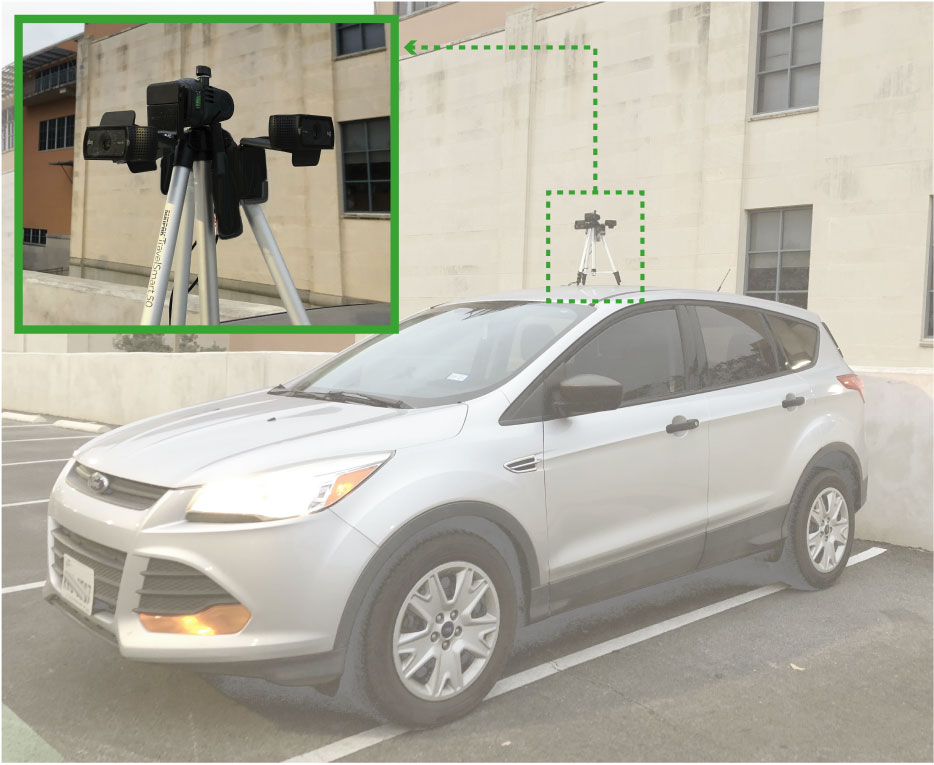}
  \caption{Field Test Platform. A Ford Escape is used with three Logitech C920 Pro Webcam cameras mounted on top of the roof facing toward the front, left, and right.}
  \label{Figure 6}
\end{figure}

\indent In training, we use Google Street View and collect 44,736 image pairs (7,456 geo-locations) of Chicago downtown areas as our training dataset. We choose Chicago because it is one of the biggest cities in the USA and its inner-city road system is very complex which makes GPS signals unreliable. Fig. \ref{Figure 5} demonstrates some target streets which are marked green. To include street scenes with diverse city settings, we purposefully select streets which include dense districts, crossroads, railway bridges, tunnels, and so forth. We employ a vehicle with three cameras mounted on the front, left, and right directions to conduct field tests to evaluate the proposed method (as shown in Fig. \ref{Figure 6}). Our experimental setting endeavors to simulate the platform of the original Google Street View data collection. We drive the vehicle to repeat approximately the same routes as the Google Street View car to collect the testing data. Our testing dataset includes 18,149 image sets, which can be converted into 60 ten-second video clips.
\subsection{Implementation Details}
In data collection, we have 3 pairs of images for one location. Each pair of images from one location is considered as a positive image pair, while a negative image pair would include images from two different locations. We use the positive and negative image pairs to train feature descriptors \cite{vo2017revisiting}. Training images are partitioned into positive and negative pairs to calculate the loss in Eq. \ref{Equation 4}. Following the implementation from \cite{dusmanu2019d2}, we perform 50 iterations using Adam Optimizer \cite{kingma2014adam}. In training, we employ a learning rate of $10^{-3}$ and decay rate of $10^{-4}$ for every 5 iterations. In addition, for image retrieval, we use different adaptive weights $\alpha$ to examine the impact on query results. By default, we use $\alpha=0.4$. 
% \textcolor{red}{Our workstation is an Intel Core i7-7820X CPU with one NVIDIA GeForce GTX 1080Ti GPU, where both training and evaluation are performed.}
\subsection{Evaluation Metric}
\indent Following\cite{sattler2018benchmarking}\cite{salarian2018improved}\cite{liu2019stochastic}, we use the $Recall@N$ of top retrieval to evaluate feature matching and image retrieval. Specifically, we count a prediction as true positive if one of the top $N$ retrievals from our location library is located within 10 meters of the ground truth of the query position. In our evaluation, we set the top $N$ as 5, 10, and 20 respectively and report the recall values associated with each different $N$. The ground truth is calculated using DJI D-RTK.\\
\indent In addition, we examine the applicability of our method for real-time vehicular operation. To achieve this goal, we examine inference accuracy and runtime. For accuracy, we use the mean Average Precision (mAP) \cite{liu2019stochastic}. We use 10-, 15-, and 20-meter thresholds to determine the true positive result. Namely, a prediction is considered to be a true positive account if the GPS of the retrieved image is located within the designated threshold distance to the ground truth of the query position. In addition, we use a per-frame inference speed to evaluate runtime performance \cite{song2019ug}.
\section{Results}
\subsection{Quantitative Analytics}
\begin{figure*}[!ht]
  \subfigure[]{
  \includegraphics[width=8cm]{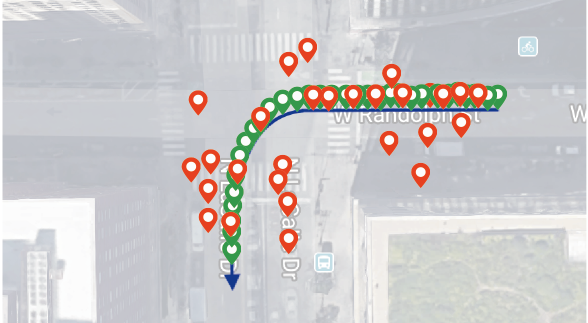}
  }
  \subfigure[]{
  \includegraphics[width=8cm]{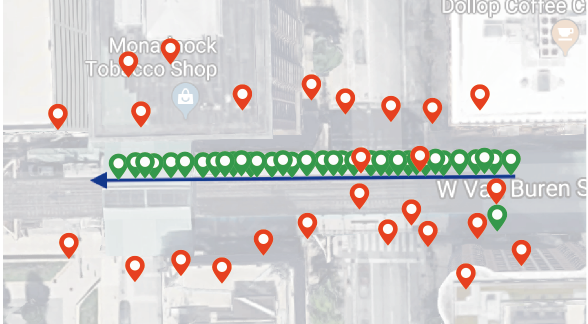}
  }  
  
  \subfigure[]{
  \includegraphics[width=8cm]{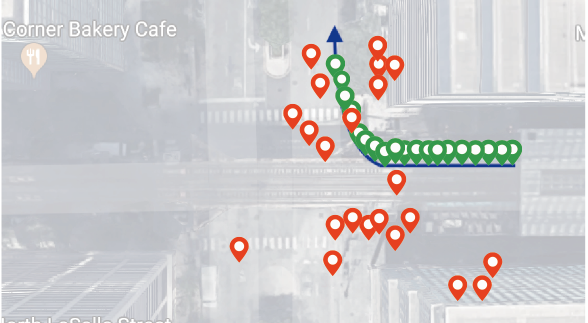}
  }  
  \subfigure[]{
  \includegraphics[width=8cm]{IMG/6-d.png}
  }
  \caption{Qualitative examples. The blue arrows and lines indicate the vehicular trajectory, the green icons are the locations predicted by the visual localization network, and the red icons show the GPS coordinates of the vehicle's assumed position. In this figure, we only demonstrate the results from the D2-Net which has the best performance among all the networks implemented in our experiment. For a clear visualization, GPS and location prediction are shown for every second.}
  \label{Figure 7}
\end{figure*}
\indent The results for $Recall@N$ performance are summarized in Table I. The results indicate that using three directional views can significantly improve localization performance, compared to those methods that only use the conventional front view. The recall rate for each of the three-directional-based methods for the same top $N$ retrieval category are all higher than their counterparts. For the three-directional-based methods, their recall values increase from 92.5\% and above at top 5 retrieval, coming close to a perfect value at top 20 retrieval. 
\begin{table}[]
\scriptsize
\centering
\caption{$Recall@N$-number of top retrieval for the different methods. $Recall@5$, $Recall@10$, and $Recall@20$ indicate top 5, top 10, and top 20 retrieval respectively. All values are reported in \%.}
\begin{tabular}{c|c|c|c|c}
\toprule
\multicolumn{2}{c|}{Method} & Recall@5 & Recall@10 & Recall@20 \\
\midrule
\multirow{2} *{D2-Net} & Front View & 58.2 & 68.1 & 70.2\\
& \textbf{Three Dimensions} & \textbf{94.6} & \textbf{96.9} & \textbf{98.2}\\
\midrule
\multirow{2}*{NetVLAD} & Front View & 50.9 & 56.8 & 60.8\\
& \textbf{Three Dimensions}  & \textbf{92.5} & \textbf{94.5} & \textbf{96.1}\\
\midrule
\multirow{2}*{SARE} & Front View & 54.7 & 60.8 & 62.3\\
& \textbf{Three Dimensions} & \textbf{93.6} & \textbf{95.5} & \textbf{96.9}\\
\bottomrule
\end{tabular}
% \caption{$Recall@N$-Number of top retrieval for the different methods. $Renall@5$, $Recall@10$, and $Recall@20$ indicate top 5, top 10 and top 20 retrieval respectively. All values are reporte,d in \%.}
\label{Table 1}
\end{table}
\indent The inference mAP and speed for each method are shown in Table II. The three-directional-view-based methods outperform the front-view-based methods under different distance thresholds pertaining to prediction accuracy. We observe that for the same method, the three-directional views improve the location predictions immediately with an affordable speed trade-off. For instance, D2-Net using three-directional views achieves 91.4\%, 95.1\%, and 98.3\% mAP under each distance threshold with a per-frame speed of 42.6 $ms$; in contrast, D2-Net using the front view achieves an inference speed of 38.2$ms$ which is a little faster than its counterpart, but the prediction accuracy for this method is 26.5\%, 24.3\%, and 21.1\% lower than its three-directional-view-based counterpart method under each distance threshold.\\
\begin{table}[!ht]
\scriptsize
    \centering
    \caption{Inference AP and speed for the different methods.}
    \begin{tabular}{cc|c|ccc}
        \toprule
        \multicolumn{2}{c|}{\multirow{2}*{Method}} & Runtime  & \multicolumn{3}{c}{mAP}\\
        \cline{4-6}
        \rule{0pt}{10pt} & & \multicolumn{1}{c|}{(f/ms)} & \multicolumn{1}{c}{10(m)} & \multicolumn{1}{c}{15(m)} & \multicolumn{1}{c}{20(m)} \\
        % Method & Runtime & 10(m) mAP & 15(m) & 20(m) \\
        \midrule 
        \multirow{2} * {D2-Net} & \textbf{Three Dimensions} & \textbf{42.6} & \textbf{91.4} & \textbf{95.1} & \textbf{98.3}\\
        & Front View & 38.2 & 64.9 & 70..8 & 77.2\\
      \midrule 
        \multirow{2} * {NetVLAD} & \textbf{Three Dimensions} & \textbf{48.5} & \textbf{88.8} & \textbf{92.6} & \textbf{93.6}\\
        & Front View & 40.9 & 47.8 & 55.5 & 68.9\\
        \midrule 
        \multirow{2} * {SARE} & \textbf{Three Dimensions} & \textbf{40.6} & \textbf{90.2} & \textbf{93.2} & \textbf{94.2} \\
        & Front View & 34.8 & 50.1 & 59.8 & 70.3\\
        \midrule
        GPS & - & - & 86.9 & 92.4 & 94.6\\
        \bottomrule 
    \end{tabular}
    \label{Table 2}
\end{table}
\indent Compared to GPS signals, all the three-directional-view-based methods have higher mAP under the 10- and 15-meter thresholds. Although GPS signals perform better than NetVLAD and SARE under the 20-meter threshold, D2-Net using three-directional views has the best overall performance in comparison. Moreover, the proposed method is more robust in tunnels or under metro bridges where GPS signals are completed blocked. Our accuracy can be further improved by interpolating GPS-tagged references between any two nearby locations from Google Street View. The 10-meter threshold stems from the fact that two close locations from Google Street View generally have 10 to 15 meters distance. 
\begin{figure}[!ht]
  \centering
\includegraphics[width=8cm, trim=0 0 0 30, clip]{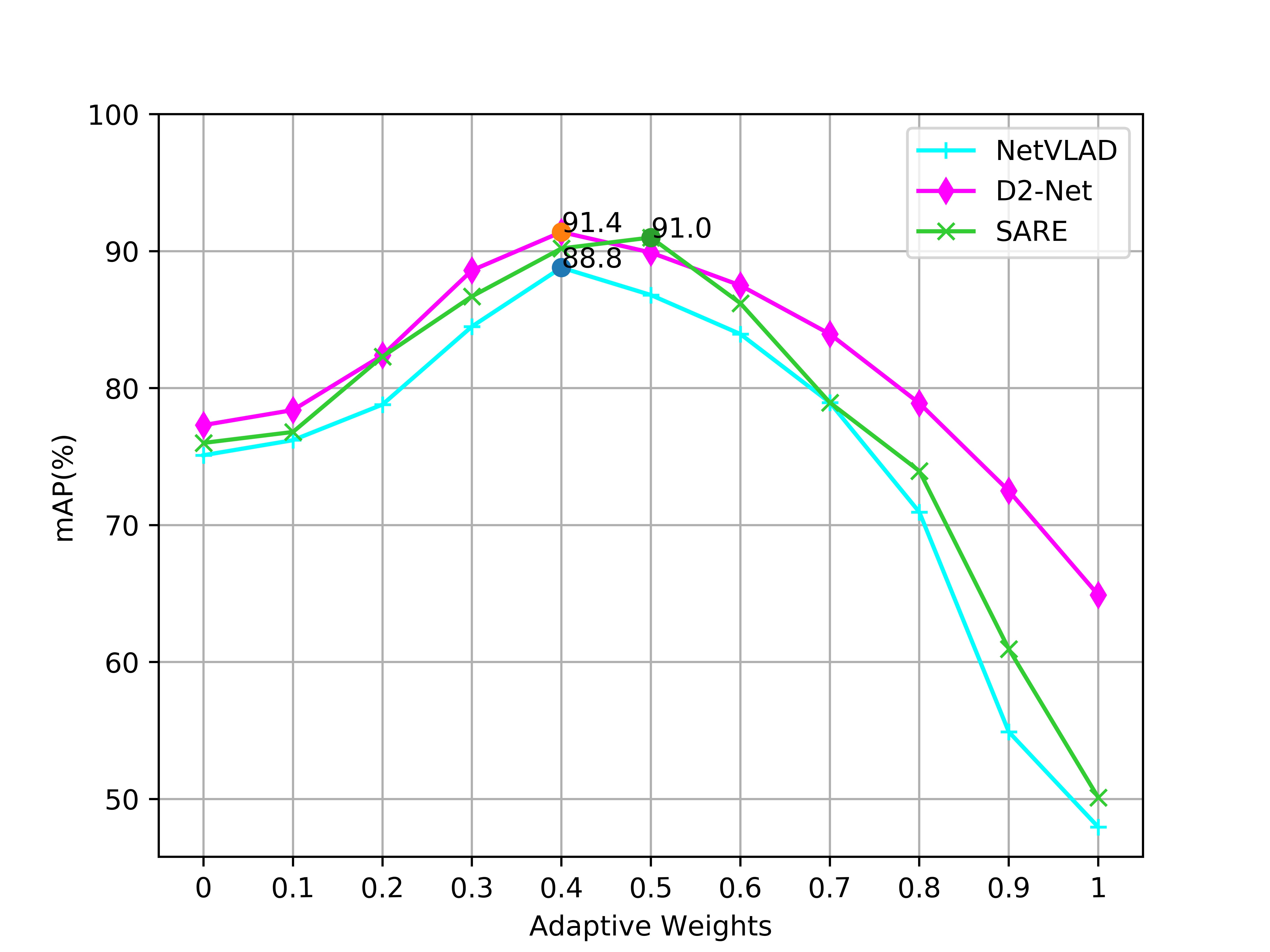}
  \caption{Adaptive weight evaluation over the whole route. We change the value of $\alpha$ from 0 to 1 which determines the contribution of the image from each direction to the final location prediction. The distance threshold is 10 meters for this plot.}
  \label{Figure 8}
\end{figure}
\subsection{Qualitative Demonstrations}
In this section, we purposefully select some examples to demonstrate the qualitative analysis (as shown in Fig. \ref{Figure 7}). Location prediction, vehicular trajectory, and GPS signals are visualized in Fig. \ref{Figure 7}. Based on our experiment, we find that GPS struggles to provide stable and reliable localization if the vehicle faces several challenging scenarios (moving by skyscrapers, under metro bridges/tunnels, or on crossroads) in an inner city. In contrast, the proposed approach can robustly predict location in the aforementioned scenarios. In Fig. \ref{Figure 7}(a) and \ref{Figure 7}(b), the vehicle is moving under a metro bridge which blocks the GPS signal significantly, while visual localization produces a reliable location inference during the process. In Fig. \ref{Figure 7}(c) and \ref{Figure 7}(d), the vehicle is moving by tall buildings (specifically, the vehicle moves under the metro before turning right or left). Under this circumstance, the location error (distance to the ground truth) is also critical for GPS signals while the predictions from our approach all fall along the correct trajectory.
\subsection{Adaptive Weight Evaluation}
\subsubsection{Over the whole route} We change the adaptive weight $a$ and examine the impact of its variations on prediction accuracy (as shown in Fig. \ref{Figure 8}). The results demonstrate that with the decrease of $\alpha$ from 1 to 0.4 (0.5 for SARE), AP for all methods increases steadily. Since the decrease of $a$ indicates that location prediction relies increasingly on left and right views, we have reasonable ground to state that adding the right and left views for localization is effective for improving the accuracy of visual localization. Then, while $a$ drops closer to 0, the accuracy for all methods starts to decrease. However, we observe that when only using views from the left and right sides ($\alpha=0$), accuracy is still higher than when only using front views ($\alpha=1$).
\begin{figure}[!ht]
  \centering
\includegraphics[width=9cm]{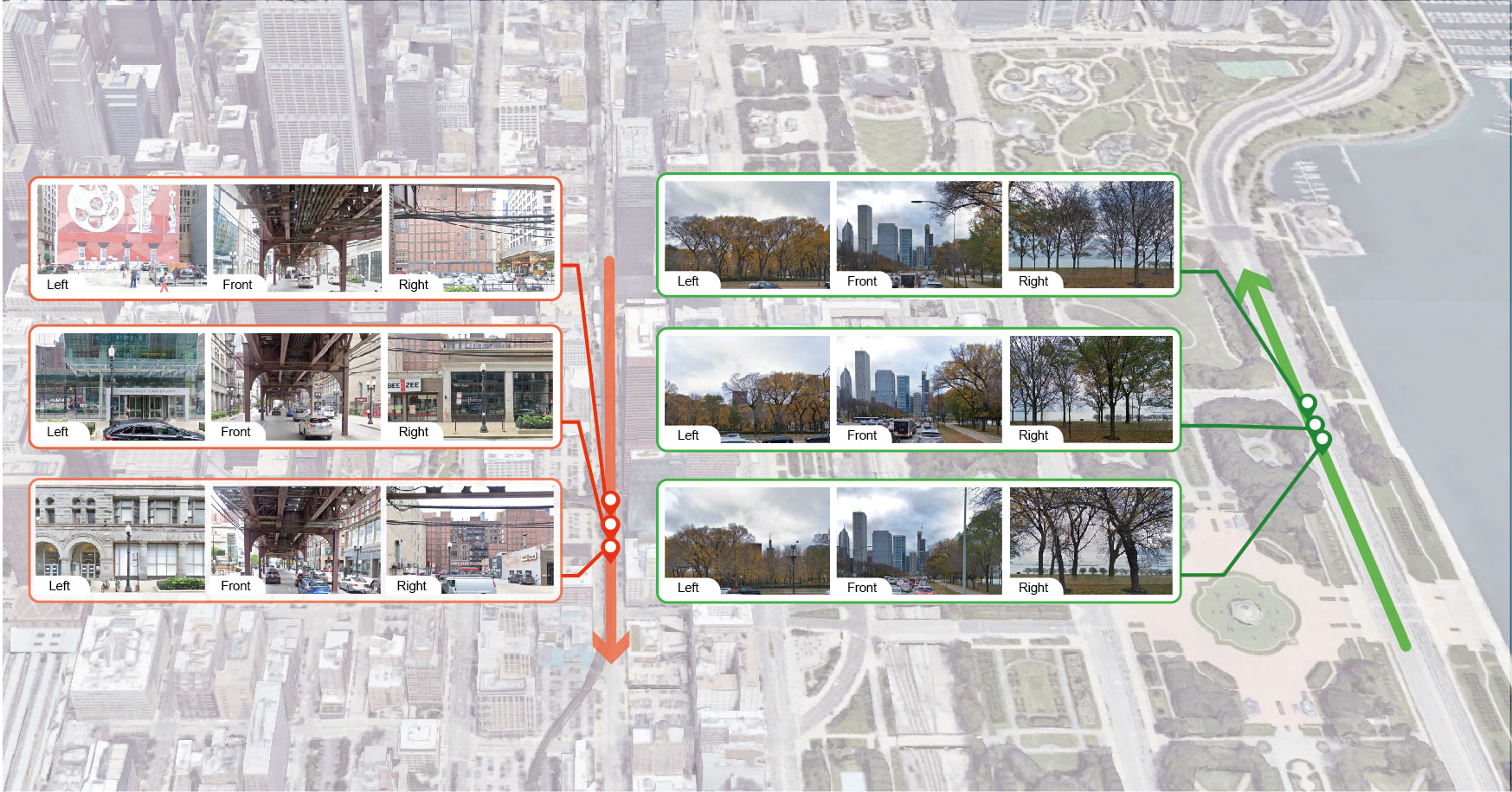}
\caption{Fast and slow street scenes. The red line and green line indicate the fast scene and slow scene respectively.}
  \label{Figure 9}
\end{figure}
\subsubsection{Under different scenarios} 
\begin{figure}[!ht]
  \centering
\includegraphics[width=7cm, trim=0 0 0 30, clip]{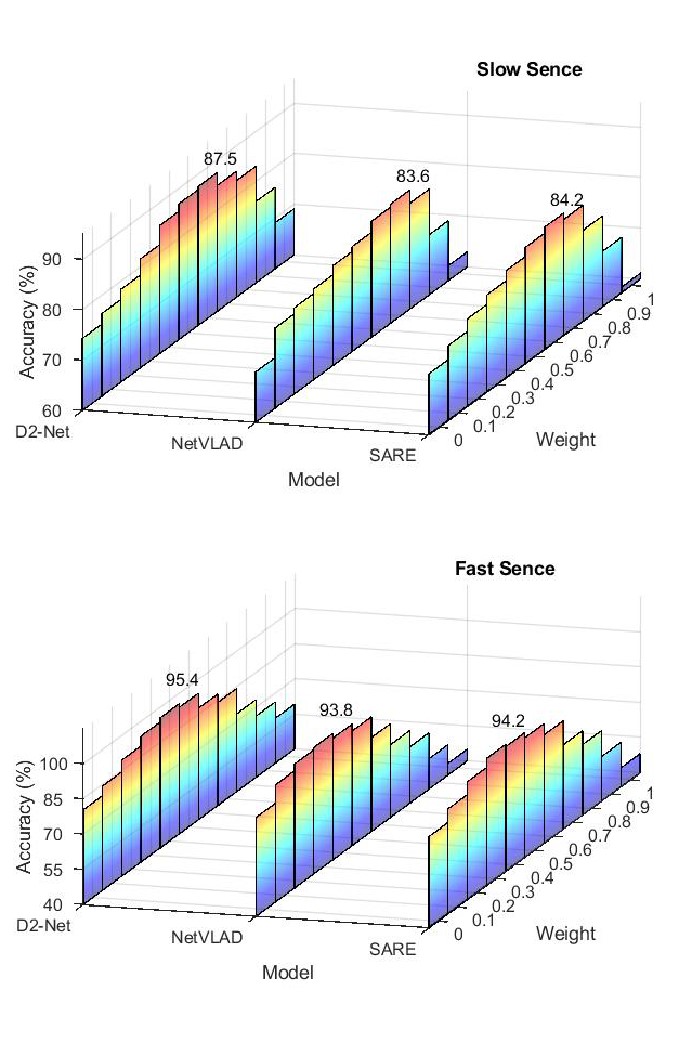}
\vskip -20pt
\caption{The results of different adaptive weights for fast and slow street scenes. The top figure is for the slow scene while the bottom figure is for the fast scene.}
  \label{Figure 10}
\end{figure}

In our experiment, we observe that the same adaptive weight could lead to different levels of prediction accuracy under different street scenes. We use motion IoU to define different street scenes based on the speed of the moving landscape caused by the movement of the vehicle. Motion IoU measures the average intersection-over-union (IoU) scores of a scene with its corresponding nearby frames (±10 frames). Our street scenes are divided into these categories: 1. slow scenes (scoring $\geq$ 0.7) in open space without any nearby buildings and 2. fast scenes (scoring $<$ 0.7) which have tall buildings along the streets. Examples from the two types of street scenes are shown in Fig. \ref{Figure 9}. Fig. \ref{Figure 10} demonstrates that the adaptive weight is sensitive to different street scenes. Under the fast street scene, left and/or right views provide more valuable information for location prediction; for the slow street scene, the front view is more informative for locating a position. Using grid search, we find that visual location networks perform best in fast scenes by using an $\alpha$ around 0.3 and in slow scenes by using an $\alpha$ around 0.6. Under different street scenes, we need to consider using different adaptive weights to facilitate robust prediction.
\section{Conclusion}
In this study, we introduce a novel solution for visual localization in inner-city areas where GPS signals are partially or completely blocked. We articulate the pipeline to implement our proposed method which uses three directional views to triangulate location. Specifically, we design an automated pipeline to collect and manage GPS-tag data from Google Street View to build a location library. We introduce a feature voting technique to determine the contributions of each directional view on location prediction. Extensive experiments demonstrate that our three-directional-based method can significantly improve the location prediction accuracy and the proposed approach for visual localization is applicable for autonomous driving.\\ \indent However, the proposed method encounters two limitations. First, the location library is built based on Google Street View, so any change of landscape would cause errors in prediction because the query image may struggle to find a matching reference in the location library. In addition, although images from Google Street View are usually obtained on days with good weather and clear visibility, we need to make our method work under adverse weather conditions, including rain, snow, fog, and even during a simple drive at night. These questions remain open for further exploration.

{\small\bibliography{test}}

\end{document}